\documentclass[11pt]{article}

\usepackage[margin=1in]{geometry}
\usepackage{amsmath, amssymb, amsthm}
\usepackage{graphicx}
\usepackage{hyperref}
\usepackage{xcolor}

\newtheorem{theorem}{Theorem}

\DeclareMathOperator*{\E}{\mathbb{E}}
\DeclareMathOperator*{\clip}{CLIP}
\DeclareMathOperator*{\mean}{mean}
\DeclareMathOperator*{\std}{std}
\DeclareMathOperator*{\Var}{Var}

\newcommand{\thetanew}{\theta}
\newcommand{\thetaold}{\theta_{\text{old}}}
\newcommand{\pion}{\pi_{\thetanew}}
\newcommand{\piold}{\pi_{\thetaold}}
\newcommand{\piphi}{\pi_{\phi}}

\newcommand{\jon}{J_{\text{on}}}
\newcommand{\joff}{J_{\text{off}}}
\newcommand{\jmix}{J_{\text{mix}}}

\newcommand{\non}{N_{\text{on}}}
\newcommand{\noff}{N_{\text{off}}}
\newcommand{\nzero}{N_{\text{zero}}}

\title{\bf Improving DAPO from a Mixed-Policy Perspective}

\author{Hongze Tan$^{1}$\hspace{0.5em}
 Yuchen Li$^{2}$ \\
  $^1$HKUST \hspace{0.5em}$^2$THU\\ 
  }

\date{}

\begin{document}

\maketitle

\begin{abstract}
This paper introduces two novel modifications to the Dynamic sAmpling Policy Optimization (DAPO) algorithm [1], approached from a mixed-policy perspective. Standard policy gradient methods can suffer from instability and sample inefficiency, particularly in sparse reward settings. To address this, we first propose a method that incorporates a pre-trained, stable guiding policy ($\piphi$) to provide off-policy experience, thereby regularizing the training of the target policy ($\pion$). This approach improves training stability and convergence speed by adaptively adjusting the learning step size. Secondly, we extend this idea to re-utilize zero-reward samples, which are often discarded by dynamic sampling strategies like DAPO's. By treating these samples as a distinct batch guided by the expert policy, we further enhance sample efficiency. We provide a theoretical analysis for both methods, demonstrating that their objective functions converge to the optimal solution within the established theoretical framework of reinforcement learning. The proposed mixed-policy framework effectively balances exploration and exploitation, promising more stable and efficient policy optimization.
\end{abstract}

\section{Introduction}

Policy gradient methods are a cornerstone of modern reinforcement learning. However, on-policy algorithms like vanilla Policy Gradient often exhibit high variance and sample inefficiency. Methods like Proximal Policy Optimization (PPO) [2] and Generational Reinforcement Policy Optimization (GRPO) [3] mitigate this by constraining the update step, ensuring the new policy does not deviate excessively from the old one. Yet, they still require the policies to be very close.

The Dynamic sAmpling Policy Optimization (DAPO) [1] algorithm introduces a unique approach but can be inefficient, especially in early training phases when the policy is not yet well-trained. The samples collected can be of low quality. Furthermore, DAPO's dynamic sampling may discard a significant number of zero-reward samples, which could contain valuable exploratory information.

In this work, we propose to enhance DAPO by introducing a mixed-policy training paradigm. Our key idea is to leverage a well-trained guiding policy, $\piphi$, which can be ``not very close" to the policy being trained, $\pion$, but is bounded by the importance sampling weights. This guiding policy provides stable, high-quality experience to accelerate and stabilize training.

We present two modifications based on this idea:
\begin{enumerate}
    \item A mixed-policy method that combines on-policy samples from $\pion$ with off-policy samples from a guiding policy $\piphi$.
    \item An extension that re-incorporates the zero-reward samples, previously discarded by DAPO, by treating them as a third set of data guided by $\piphi$.
\end{enumerate}
We demonstrate theoretically that these modifications are well-founded, leading to convergent algorithms that find theoretically optimal solutions.

\section{Theoretical Foundation}

Our analysis builds upon the convergence properties of policy gradient algorithms. We start with a foundational theorem [4].

\begin{theorem}
Suppose the objective function of the policy gradient algorithm $J \in \mathcal{J}_n$, where $\mathcal{J}_n$ is the class of finite-sum Lipschitz smooth functions, has $\sigma$-bounded gradients, and the importance weight $w = \pion / \piphi$ is clipped to be bounded by $[\underline{w}, \bar{w}]$. Let the learning rate $\alpha_k = \alpha = c/\sqrt{K}$, where
\[
c = \sqrt{\frac{2(J(\theta^*) - J(\theta^0))}{L \sigma^2 \bar{w}}},
\]
and $\theta^*$ is an optimal solution. Then, the iterates of our algorithm of $J(\theta)$ satisfy
\[
\min_{0 \le k \le K-1} \E \left[ \|\nabla J(\theta^k)\|^2 \right] \le \sqrt{\frac{2(J(\theta^*) - J(\theta^0)) L\bar{w}}{K\underline{w}}} \sigma.
\]
\end{theorem}
\noindent This theorem provides a convergence guarantee for policy gradient algorithms under certain conditions, which forms the basis for our proposed methods.

\section{Proposed Modifications}

We introduce two modifications to the DAPO algorithm from a mixed-policy perspective.

\subsection{Method 1: Mixed Policy with Off-policy Guidance}

In the initial stages of training, the target policy $\pion$ is often suboptimal. To improve sample efficiency, we introduce a well-trained guiding policy $\piphi$ to ``guide" the training of $\pion$. Unlike PPO, $\pion$ and $\piphi$ do not need to be very close, as long as their ratio is bounded as per Theorem 1.

\begin{itemize}
    \item \textbf{On-policy Sampling:} We first generate $\non$ sequences of samples using the current policy $\pion$. Let this set be $\{u_i\}_{i=1}^{\non}$.
    \item \textbf{Off-policy Sampling:} We then generate $\noff$ sequences using the guiding policy $\piphi$. Let this set be $\{v_j\}_{j=1}^{\noff}$.
\end{itemize}

Following DAPO, the reward for any token within a sequence is defined as the final reward of that sequence. Let $R(u_i)$ and $R(v_j)$ be the rewards. We define two separate advantage functions based on this principle:
\begin{equation}
    \hat{A}_{i,t} = \hat{A}_i = \frac{R(u_i) - \mean(\{R(u_k)\}_{k=1}^{\non})}{\std(\{R(u_k)\}_{k=1}^{\non})}
\end{equation}
\begin{equation}
    \hat{B}_{j,t} = \hat{B}_j = \frac{R(v_j) - \mean(\{R(v_k)\}_{k=1}^{\noff})}{\std(\{R(v_k)\}_{k=1}^{\noff})}
\end{equation}

Next, we define the objective functions. The on-policy objective is:
\begin{equation}
    \jon(\theta) = \frac{1}{\sum_{i=1}^{\non}|u_i|} \sum_{i=1}^{\non}\sum_{t=1}^{|u_i|} \clip(r^{(1)}_{i,t}(\theta), \hat{A}_i, \epsilon), \quad \text{where} \quad r^{(1)}_{i,t}(\theta) = \frac{\pion(u_{i,t}|q, u_{i,<t})}{\piold(u_{i,t}|q, u_{i,<t})}.
\end{equation}
The off-policy objective, guided by $\piphi$, is:
\begin{equation}
    \joff(\theta) = \frac{1}{\sum_{j=1}^{\noff}|v_j|} \sum_{j=1}^{\noff}\sum_{t=1}^{|v_j|} f(\hat{r}_{j,t}(\theta, \phi)) \cdot \hat{B}_j,
\end{equation}
where $f(x) = \frac{x}{x+\gamma}$ is a scaling function with a small constant $0 < \gamma \ll 1$, and be careful $\hat{r}_{j,t}(\theta, \phi) = \frac{\pion(v_{j,t}|q, v_{j,<t})}{\piphi(v_{j,t}|q, v_{j,<t})}$.

The final objective function combines both parts. Assuming $\non \approx \noff$, we can use equal weights:
\begin{equation} \label{eq:J1}
    J_{\text{Mix\_1}}(\theta) = \frac{1}{2}\jon(\theta) + \frac{1}{2}\joff(\theta).
\end{equation}

\subsubsection{Gradient Analysis and Convergence}
The gradient of the off-policy objective is:
\[
\nabla_\theta \joff(\theta) = \E_{v \sim \piphi} \left[ f'\left(\frac{\pion(v)}{\piphi(v)}\right) \frac{\pion(v)}{\piphi(v)} \nabla_\theta \log \pion(v) \cdot \hat{B}_v \right].
\]
The gradient of the on-policy objective is the standard importance-sampling gradient estimator:
\[
\nabla_\theta \jon(\theta) = \E_{u \sim \piold} \left[ \frac{\pion(u)}{\piold(u)} \nabla_\theta \log \pion(u) \cdot \hat{A}_u \right].
\]
Since $f'(x) = \frac{\gamma}{(x+\gamma)^2} > 0$, the gradient of $\joff(\theta)$ has a similar form to that of $\jon(\theta)$, differing only by a positive scaling factor. Based on Theorem 1, $\joff(\theta)$ will converge. If we use the same optimization algorithm for both gradients, they will both approach their respective theoretical optima. This implies:
\[
\nabla_\theta J_{\text{Mix\_1}}(\theta) = 0 \iff \nabla_\theta \joff(\theta) = 0 \iff \nabla_\theta \jon(\theta) = 0.
\]
This result guarantees that the extremum we find by optimizing $J_{\text{Mix\_1}}(\theta)$ is a valid optimum in the reinforcement learning sense.

\subsubsection{Benefits of the Modification}
\begin{itemize}
    \item \textbf{Adaptive Step Size:} In early training, when $\pion$ is far from producing correct answers, the ratio $\pion/\piphi$ is small. This makes $f'(\pion/\piphi) \approx 1/\gamma \gg 1$, leading to a larger gradient step for the off-policy component. This accelerates learning by strongly guiding $\theta$ towards the expert policy's behavior.
    \item \textbf{Variance Reduction:} As training progresses and $\pion$ approaches $\piphi$, the ratio $\pion/\piphi$ approaches 1. In this regime, $f'(\pion/\piphi) \approx \gamma$ is small. The variance of the off-policy gradient, $\Var(\nabla_\theta \joff)$, becomes approximately $\gamma^2 \Var(\nabla_\theta J_{\text{Mix\_1}}^*)$, where $\nabla_\theta J_{\text{Mix\_1}}^*$ is the traditional off-policy gradient. The scaling function $f(\cdot)$ thus stabilizes training by reducing gradient variance when the policies are close.
\end{itemize}

\subsection{Method 2: Reusing Zero-Reward Samples}

DAPO's dynamic sampling discards trajectories with zero reward, which can be wasteful. We propose to utilize these samples by treating them as a separate batch guided by the expert policy $\piphi$.

The sampling process is now threefold:
\begin{itemize}
    \item \textbf{On-policy (non-zero reward):} $\non$ samples $\{u_i\}_{i=1}^{\non}$ from $\pion$ with $R(u_i) > 0$.
    \item \textbf{On-policy (zero reward):} $\nzero$ samples $\{w_k\}_{k=1}^{\nzero}$ from $\pion$ with $R(w_k) = 0$.
    \item \textbf{Off-policy:} $\noff$ samples $\{v_j\}_{j=1}^{\noff}$ from the guiding policy $\piphi$.
\end{itemize}
We define three corresponding advantage functions. $\hat{A}_i$ is the same as before. $\hat{B}_j$ and $\hat{C}_k$ are now normalized over the combined pool of off-policy and zero-reward samples to share a common baseline:
\begin{equation}
    \hat{B}_{j,t} = \hat{B}_j = \frac{R(v_j) - \mean(\{R(v)\}_{v \in V \cup W})}{\std(\{R(v)\}_{v \in V \cup W})}
\end{equation}
\begin{equation}
    \hat{C}_{k,t} = \hat{C}_k = \frac{R(w_k) - \mean(\{R(v)\}_{v \in V \cup W})}{\std(\{R(v)\}_{v \in V \cup W})}
\end{equation}
where $V = \{v_j\}_{j=1}^{\noff}$ and $W = \{w_k\}_{k=1}^{\nzero}$.

The on-policy objective $\jon(\theta)$ remains the same. We introduce a new mixed objective $\jmix(\theta)$ that combines the zero-reward and off-policy samples:
\begin{equation}
    \jmix(\theta) = \frac{1}{\sum_{j=1}^{N_{off}}|v_j|+\sum_{k=1}^{N_{zero}}|w_k|} \bigg(\underbrace{\sum_{j,t} f(\hat{r}_{j,t}(\theta,\phi))\cdot \hat{B}_j}_{\text{off-policy part}} + \underbrace{\sum_{k,t} \clip(r^{(2)}_{k,t}(\theta), \hat{C}_k, \epsilon)}_{\text{zero-reward part}}\bigg)
\end{equation}
where $r^{(2)}_{k,t}(\theta) = \frac{\pion(w_{k,t}|q, w_{k,<t})}{\piphi(w_{k,t}|q, w_{k,<t})}$. Note the importance sampling for the zero-reward samples is also done with respect to $\piphi$.

The final objective is a combination of the on-policy and mixed objectives:
\begin{equation}
    J_{\text{Mix\_2}}(\theta) = \frac{1}{2} \jon(\theta) + \frac{1}{2} \jmix(\theta).
\end{equation}
The gradient of the mixed term is a weighted sum:
\begin{equation}
    \nabla_\theta \jmix(\theta) = l_1 \cdot \nabla_\theta J_1(\theta) + l_2 \cdot \nabla_\theta J_2(\theta),
\end{equation}
where $\nabla_\theta J_1(\theta)$ corresponds to the off-policy gradient and $\nabla_\theta J_2(\theta)$ corresponds to the zero-reward sample gradient, with $l_1+l_2=1$.

Similar to the first method, the analysis holds. The gradients for all three components ($\jon, J_1, J_2$) are structurally similar up to positive scaling factors. Therefore, optimizing the composite objective $J_{\text{Mix\_2}}(\theta)$ will drive the gradients of all components to zero, ensuring convergence to a theoretically sound optimum.

\section{Conclusion}

This paper presented two extensions to the DAPO algorithm based on a mixed-policy framework. By incorporating a stable guiding policy, we can improve the stability and efficiency of training. The first method uses off-policy samples to regularize learning, while the second method further enhances sample efficiency by re-utilizing zero-reward samples that are typically discarded.

Our theoretical analysis confirms that these modifications are well-founded and lead to convergent algorithms that find optimal policies. These theoretically-motivated improvements balance exploration with stable exploitation, addressing key challenges in policy gradient methods. Future work will involve empirical validation of these methods on relevant benchmarks to quantify their performance gains.


\end{document}